\documentclass[9pt]{article}
\usepackage{spconf,amsmath,graphicx}

\usepackage{graphicx}
\usepackage{subcaption}
\usepackage{amsmath}
\usepackage{algorithm}
\usepackage{algorithmic}
\usepackage{booktabs}
\usepackage{float}

\title{Absolute 3D Pose Estimation and Length Measurement of Severely Deformed Fish from Monocular Videos in Longline Fishing}
\name{Jie Mei$^{\star}$\thanks{$^{\star}$e-mail:\{jiemei,hwang\}@uw.edu}, Jenq-Neng Hwang$^{\star}$, Suzanne Romain$^{\dagger}$, Craig Rose$^{\dagger}$, Braden Moore$^{\dagger}$, Kelsey Magrane$^{\dagger}$\thanks{$^{\dagger}$e-mail:\{suzanne.romain,craig.rose,braden.j.moore,kelsey.magrane\}
@noaa.gov}}
  
\address{$^{\star}$ Department of ECE,
University of Washington,
Seattle, WA, USA\\
$^{\dagger}$Pacific States Marine Fisheries Commission, National Oceanic and Atmospheric Administration, USA}

\begin{document}

%
\maketitle
\begin{abstract}
Monocular absolute 3D fish pose estimation allows for efficient fish length measurement in the longline fisheries, where fishes are under severe deformation during the catching process. This task is challenging since it requires locating absolute 3D fish keypoints based on a short monocular video clip. Unlike related works, which either require expensive 3D ground-truth data and/or multiple-view images to provide depth information, or are limited to rigid objects, we propose a novel frame-based method to estimate the absolute 3D fish pose and fish length from a single-view 2D segmentation mask. We first introduce a relative 3D fish template. By minimizing an objective function, our method systematically estimates the relative 3D pose of the target fish and fish 2D keypoints in the image. Finally, with a closed-form solution, the relative 3D fish pose can help locate absolute 3D keypoints, resulting in the frame-based absolute fish length measurement, which is further refined based on the statistical temporal inference for the optimal fish length measurement from the video clip. Our experiments show that this method can accurately estimate the absolute 3D fish pose and further measure the absolute length, even outperforming the state-of-the-art multi-view method.
\end{abstract}
\section{Introduction}
\label{sec:intro}
Given one single image of a deformed fish such as Pacific Halibut during longline fishing~\cite{mei2021videobased}, our goal is to design a pipeline that receives as input the whole single-view image and produces as outputs the absolute 3D keypoints location and length of the fish~\cite{ huang2016chute} . 

\begin{figure*}[htpb]
\centering
\includegraphics[width=0.84\textwidth]{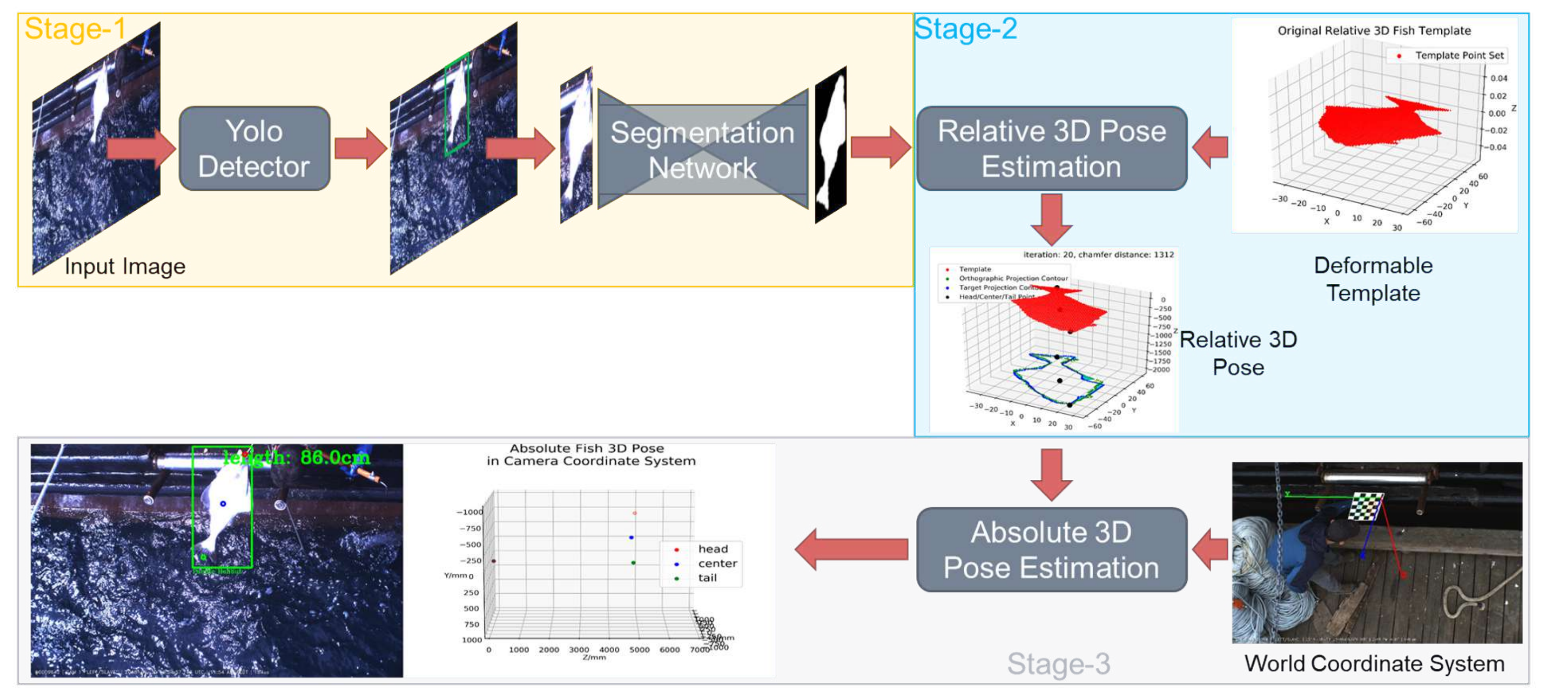}
\caption{Pipeline: Note that in the world coordinate system, the red axis is $Z$ axis. In the left bottom corner, camera image plane is on $Z=0$ plane, and $Z$ axis is depth, with unit being millimeter. So, the above fish is about 5 meters away from the camera.}
\label{pipeline}
\end{figure*}

As shown in Fig.\ref{pipeline}, in Stage-1, we use the YOLO object detector~\cite{redmon2016you} and an encoder-decoder FCN architecture for instance segmentation~\cite{ronneberger2015u}, refined by histogram back projection~\cite{chuang2011automatic}. We have ground truth labels to train both models.  

Our main contributions are in the last two stages. Stage-2 introduces a relative 3D fish template and chamfer distance loss~\cite{DBLP:journals/corr/FanSG16}. Only relying on the target's 2D segmentation mask from Stage-1, Stage-2 is able to estimate the relative 3D pose of the target and three 2D keypoints in image. Stage-3 introduces a novel 3D localization method to locate these keypoints in absolute 3D space and measure the length of the fish.

\begin{figure}
    \centering
    \begin{minipage}[c]{0.16\textwidth}
        \centering
        \includegraphics[height= 0.15\textheight]{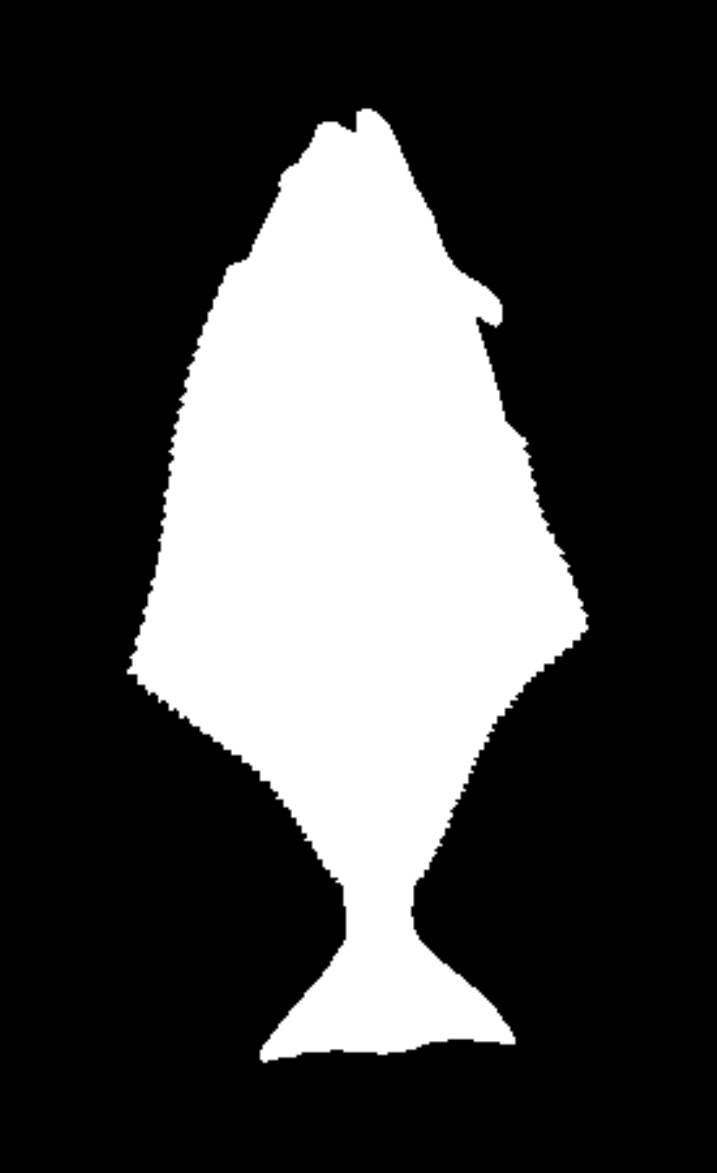}
        \subcaption{}
    \end{minipage}
    \begin{minipage}[c]{0.3\textwidth}
        \centering
        \includegraphics[height= 0.15\textheight]{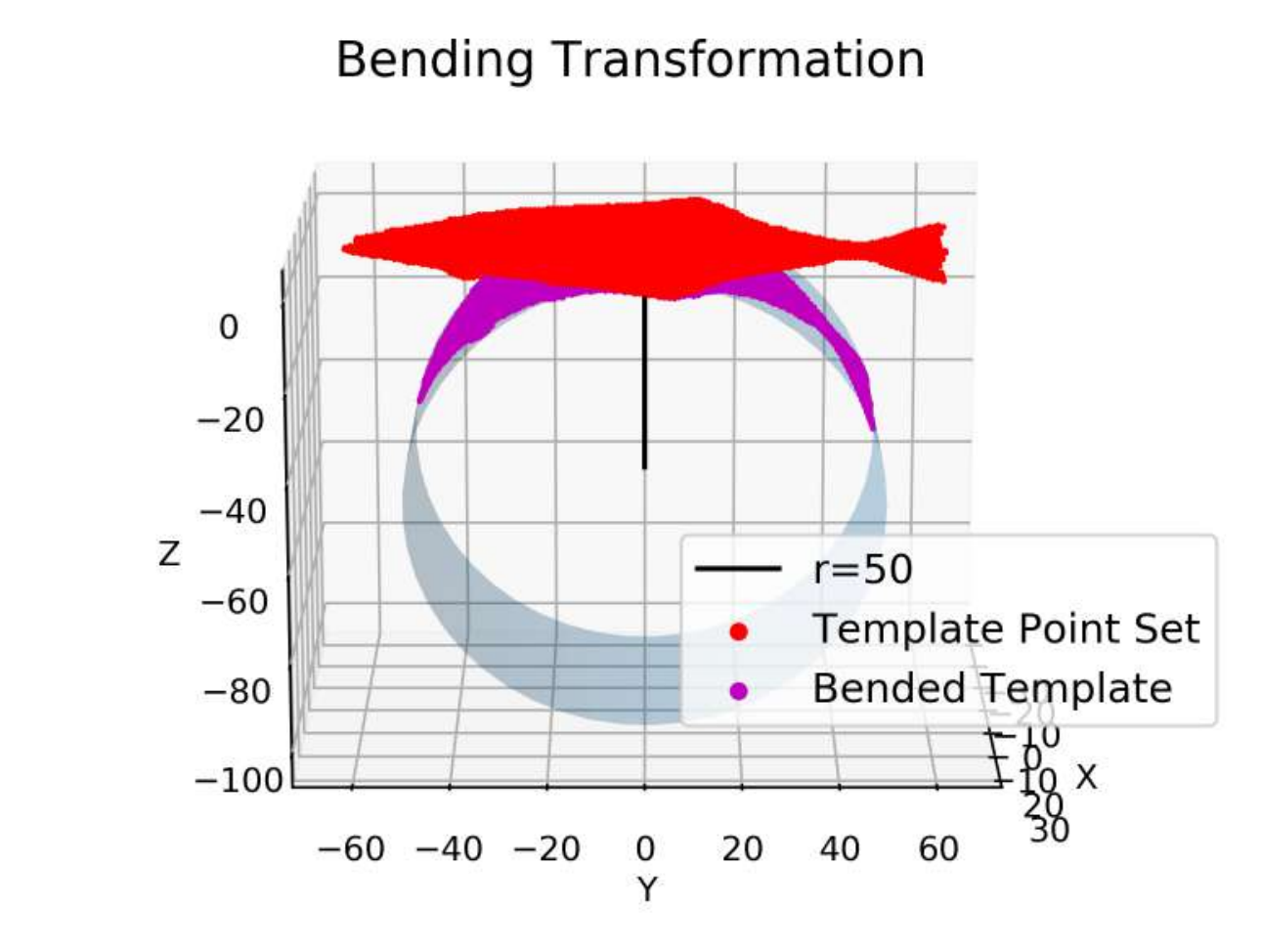}
        \subcaption{}
    \end{minipage}
    \caption{3D Template: (a) A binary mask of a standard flat Pacific Halibut; (b) The initial original 3D point set of foreground pixels is the relative 3D fish template on $Z=0$ plane.}
    \label{template}
\end{figure}

\section{Proposed Method}
\label{sec:format}
\subsection{Relative 3D Fish Template}
\label{ssec:subhead}
Inspired by~\cite{wang2019multi, NIPS2012_4562}, we introduce a relative 3D fish template, whose unit is a pixel. It is a 3D point set on a surface generated from a Pacific Halibut Template, as illustrated in Fig.\ref{template}. The origin point is the center point of the fish body. We denote the initial template point set as $S_{0} = \{(x_{0i}, y_{0i}, z_{0i}) \in R^3\}_{i=1}^{N}$, where $N$ is the total number, and $0$ represents the initial point set before any deformable transformation. To simulate the deformable pose of the fish, the template iteratively executes the following four transformation sequentially:

\textbf{Scale Transformation} We denote the scale parameter as $s \in R$. The scaling transformation is to multiply $s$ with each point, $(x_{0i}, y_{0i}, z_{0i})$, in the template point set $S_0$:
\begin{equation}
\left(x_{1i}, y_{1i}, z_{1i}\right)=s \cdot\left(x_{0i}, y_{0i}, z_{0i}\right),
\end{equation}
resulting in a new point set $S_{1} = \{(x_{1i}, y_{1i}, z_{1i}) \in R^3\}_{i=1}^{N}$.

\textbf{Bending Transformation} We denote the bending parameter as $r \in R$ being the radius of a cylinder, which is tangent with the template at the initial position, as illustrated in Fig.\ref{template}. The bending transformation maps each point, $(x_{1i}, y_{1i}, z_{1i})$, in $S_1$ to a new 3D point, $(x_{2i}, y_{2i}, z_{2i})\in R^3$, on the cylinder:
\begin{equation}
\begin{aligned}
\begin{cases}
x_{2i}&=x_{1i}, \\
z_{2i}&=r\left(1-\cos \frac{y_{1i}}{r}\right),\\
y_{2i}&=\frac{z_{2i}}{\tan \frac{y_{1i}}{2r}},\\
\end{cases}
\end{aligned}
\end{equation}
which is fish length preserving, resulting in a new point set, $S_{2} = \{(x_{2i}, y_{2i}, z_{2i}) \in R^3\}_{i=1}^{N}$ (the purple surface in Fig.\ref{template}).

\textbf{Translation Transformation} We denote the translation parameters as $(T_x, T_y, 0)\in R^3$:
\begin{equation}
\left(x_{3i}, y_{3i}, z_{3i}\right)=\left(x_{2i}, y_{2i}, z_{2i}\right) + \left(T_{x}, T_{y}, 0\right), 
\end{equation}
resulting in a new point set $S_{3} = \{(x_{3i}, y_{3i}, z_{3i}) \in R^3\}_{i=1}^{N}$.

\textbf{Rotation Transformation} We denote three rotation parameters as $\alpha, \beta, \gamma\in [0, 2\pi]$, which respectively represent the rotation degree around $x$, $y$, $z$ axis~\cite{arvo1992fast}. We construct three basic (elemental) rotation matrices, $R(\gamma)$, $R(\beta)$, and $R(\alpha) \in R^{3\times3}$ and rotation transform each point, $(x_{3i}, y_{3i}, z_{3i})$ to a new point $(x_{4i}, y_{4i}, z_{4i})\in R^3$:
\begin{equation}
\begin{aligned}
\left(x_{4i}, y_{4i}, z_{4i}\right)=\left(x_{3i}, y_{3i}, z_{3i}\right)\cdot R(\gamma) \cdot R(\beta) \cdot R(\alpha).
\end{aligned}
\end{equation}


Finally, after performing these four transformations one-by-one iteratively, we denote the final relative 3D template as $S_{4}=\{(x_{4i}, y_{4i}, z_{4i})\}_{i=1}^{N}\in R^3$. These four transformations are differentiable with respect to their parameters, which enables us to do relative 3D pose estimation in the next section.

\begin{figure*}
    \centering
    \begin{minipage}[c]{0.12\textwidth}
        \centering
        \includegraphics[height= 0.15\textheight]{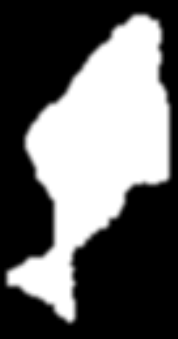}
        \subcaption{}
        \label{fig:figure-a}
    \end{minipage}
  \begin{minipage}[c]{0.12\textwidth}
        \centering
        \includegraphics[height= 0.15\textheight]{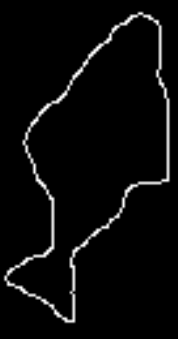}
         \subcaption{}
    \end{minipage}
    \begin{minipage}[c]{0.24\textwidth}
        \centering
        \includegraphics[height= 0.15\textheight]{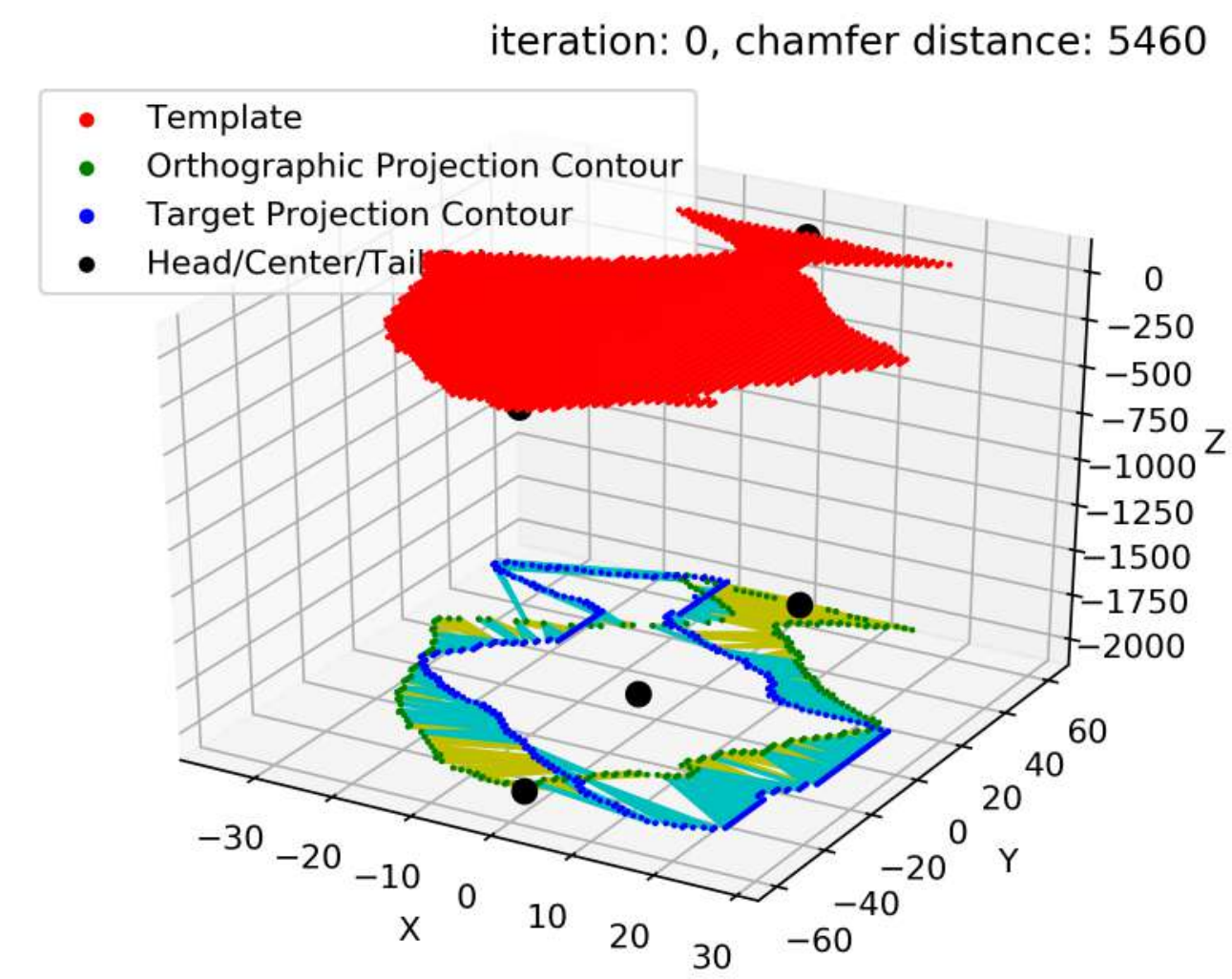}
        \subcaption{ }
    \end{minipage}
    \begin{minipage}[c]{0.24\textwidth}
        \centering
        \includegraphics[height= 0.15\textheight]{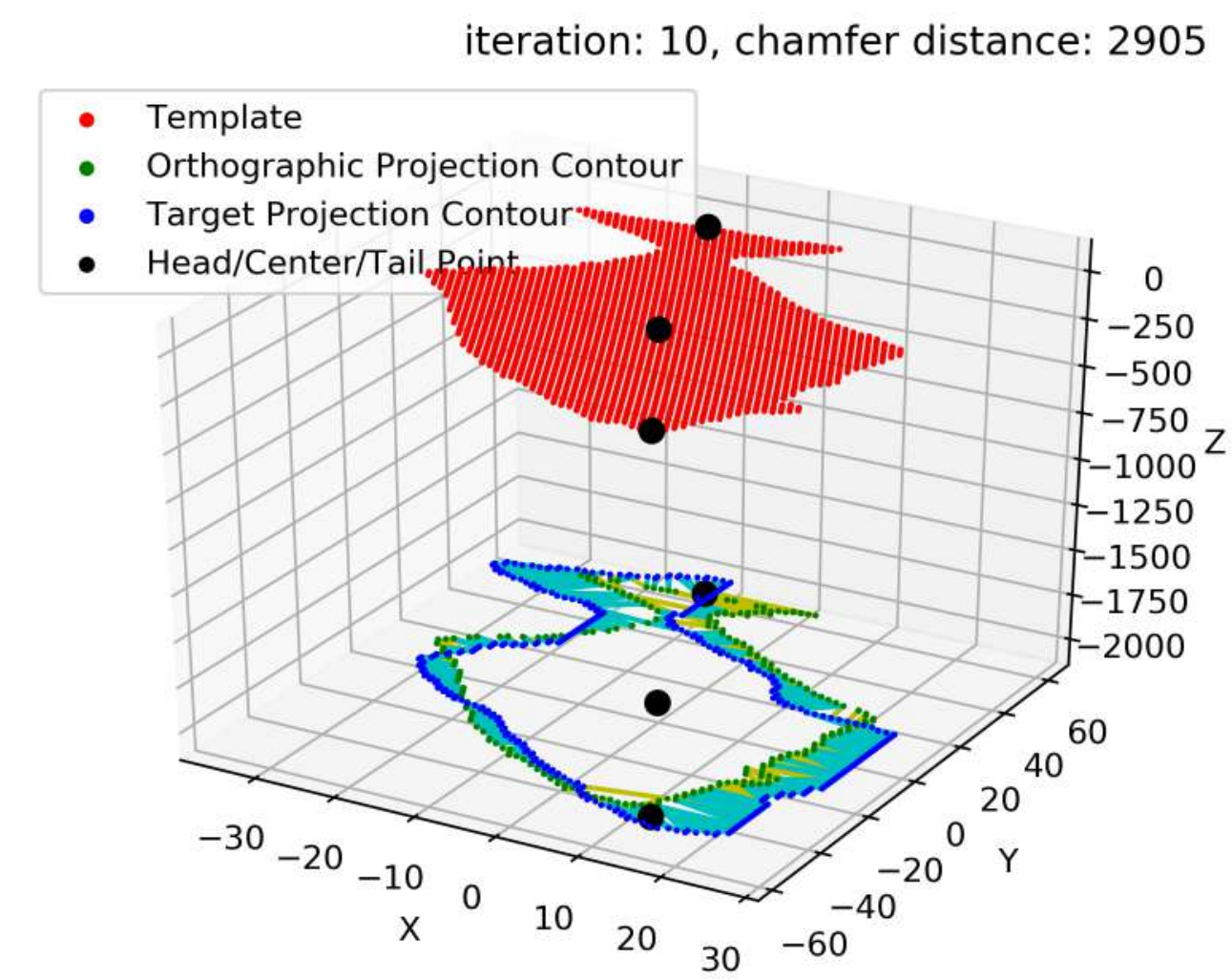}
        \subcaption{}
    \end{minipage}
    \begin{minipage}[c]{0.24\textwidth}
        \centering
        \includegraphics[height= 0.15\textheight]{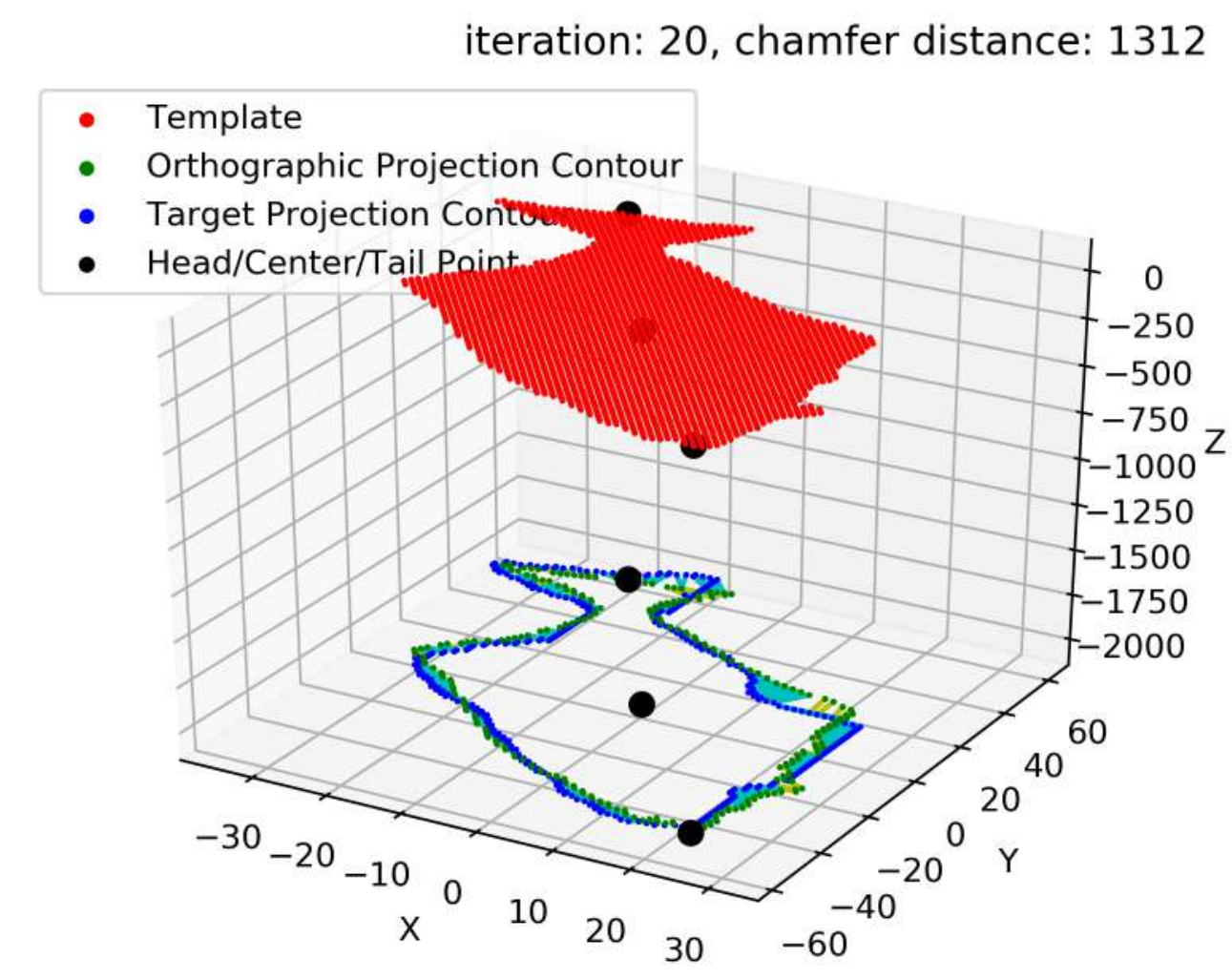}
        \subcaption{}
    \end{minipage}
    \caption{Relative 3D Pose Estimation: (a) A target mask from Stage-1; (b) After applying a Canny edge detector on (a), indicated by the blue point set in (c); (c) Black dots are predefined head/center/tail keypoints, yellow connections and cyan connections respectively are the first term and second term in chamfer distance. Those connections are ‘forces' to pull the template contour close to the target contour, a similar concept with 'intrinsic/extrinsic forces' in~\cite{terzopoulos1988constraints}; (d) A middle shot; (e) The final shot.}
    \label{optimization}
\end{figure*}

\subsection{Relative 3D Pose Estimation}
\label{ssec:subhead}
Based on a target's 2D mask, we can directly estimate the relative 3D pose of the target fish and 2D keypoints in image without using expensive 3D ground truth~\cite{wang2018pixel2mesh, DBLP:journals/corr/abs-1901-11461} or multi-view images. We iteratively minimize the chamfer distance between the orthographic projection contour of 3D template and one target's 2D mask contour, as illustrated in Fig.\ref{optimization}. 

\textbf{Orthographic Projection Contour} We get the orthographic projection points of the 3D template by directly setting each point, $(x_{4i}, y_{4i}, z_{4i})$ to $(x_{4i}, y_{4i}, 0)$. Then, we convert these points into a binary mask and execute Canny edge detector~\cite{4767851} and denote the orthographic projection contour points as, $S_{template}=\{(x_{4i}, y_{4i}, z_{4i})\in R^3\}_{i}^{n}\in S_4$, where $n$ varies given different transformations' parameters. 


\textbf{Target Contour} The target contour points are acquired by applying the Canny edge detector on the target's 2D binary mask from Stage-1. Because the template's center point is at the origin point before four transformations, we also transform the mask image center to the origin point and then get the final target contour point set, denoted as $S_{target}=\{(x_{j}, y_{j}, z_{j}) \in R^3 \}_{j=1}^{m}$.

\textbf{Chamfer Distance Loss} Given a fixed $S_{target}$, and an adjustable $S_{template}$, we iteratively and sequentially adjust each point in $S_{template}$ with seven parameters in four deformation transformations so that the chamfer distance, Eq. \ref{chamfer}, is minimized by gradient descent~\cite{DBLP:journals/corr/Ruder16}: 
\begin{small}
\begin{equation}
\begin{aligned}
d_{C D}\left(S_{template}, S_{target}\right)=\sum_{p_i \in S_{template}} \min _{q_j \in S_{target}}\|p_i-q_j\|_{2}^{2}\\
+\sum_{q_j \in S_{target}} \min _{p_i \in S_{template}}\|p_i-q_j\|_{2}^{2},
\end{aligned}
\label{chamfer}
\end{equation}
\end{small}
where the first term implies, for each point $ p_i = (x_{4i}, y_{4i}, z_{4i})\in S_{template}$, calculate the Euclidean distance of this point and its nearest  point $ q_j = (x_{j}, y_{j}, z_{j}) \in S_{target}$. The second term implies, for each point in $ q_j= (x_{j}, y_{j}, z_{j}) \in S_{target}$, find the nearest  point $ p_i=(x_{4i}, y_{4i}, z_{4i}) \in S_{template}$ to calculate their Euclidean distance. After optimization, the relative 3D fish template's pose is the estimated target's relative 3D pose in the camera coordinate system but with $pixel$ as a unit. Plus, since we can predefine three keypoints (head, center and tail) on the template, their orthographic projections are the 2D keypoints in the target mask, as illustrated in Fig.\ref{optimization}, denoted as $h_{2d}$ $(U_h,V_h)$, $c_{2d}$ $(U_c,V_c)$, $t_{2d}$ $(U_t,V_t)$ respectively in the whole image coordinate system. We denote center, head, and tail points in target's relative 3D pose as $h$ $(x_h, y_h, z_h)$, $c$ $(x_c, y_c, z_c)$, and $t$ $(x_t, y_t, z_t)$ respectively.

\subsection{Absolute 3D Localization}
To have a 3D length measurement of the fish body, we introduce an absolute 3D localization method to locate three keypoints in absolute 3D, denoted respectively as $H'$ $(X_{hc}, Y_{hc}, Z_{hc})$, $C'$ $(X_{cc}, Y_{cc}, Z_{cc})$, and $T'$ $(X_{tc}, Y_{tc}, Z_{tc})$, where the second subscript $c$ means the camera coordinate system and the unit is $millimeter$ instead of $pixel$. This task is formulated as the following closed-form solution. 

\textbf{Back Projection} We first use camera intrinsic parameters to back project $h_{2d}$ $(U_h,V_h)$, $c_{2d}$ $(U_c,V_c)$, $t_{2d}$ $(U_t,V_t)$ to 3D space in the camera coordinate system without depth:
\begin{equation}
\left[\begin{array}{c}
X \\
Y\\
1
\end{array}\right]
=K^{-1}\left[\begin{array}{c}
U \\
V \\
1
\end{array}\right],
\end{equation}
where $K$ denotes camera intrinsic parameters, that is a $3\times3$ matrix, obtained by method from~\cite{888718}. Now for $h_{2d}$, $c_{2d}$, $t_{2d}$, respectively we get $h''$ $(X_h, Y_h, 1)$, $c''$ $(X_c, Y_c, 1)$, and $t''$ $(X_t, Y_t, 1)$ in the camera coordinate system.


\textbf{Reference Plane} In the longline fishing~\cite{huang2018fish, williams2016automated}, all fish are hooked and pulled up by a line. So our assumption is that all fish's center points are always on a reference plane, which coincides with the plane of checkerboard of known grid size used in the camera calibration, defined as $Z = 0$ plane in the world coordinate system, as illustrated in the right bottom corner of Fig.\ref{pipeline}. Then we use the solvePnP method~\cite{1217599} to calculate the rotation matrix, $R_{3\times3}$, and translation vector, $T_{3\times1}$, between the world coordinate system and the camera coordinate system. Under our assumption, center point $C'$ is a point $(X_{cw}, Y_{cw}, 0)$ in the world coordinate system. With $R_{3\times3},\ T_{3\times1}$ and $K_{3\times3}$, we can get $Z_{cc}$ (depth) with Eq.\ref{Zcc}. We also need to calculate homography matrix $H$ between the image plane and $Z=0$ plane in the world coordinate system:
\begin{equation}
\begin{aligned}
\begin{cases}
H \quad = \quad K_{3\times3} & \cdot\left[\begin{array}{lll}
R_{1} & R_{2} & T
\end{array}\right]_{3\times3},\\
\left[\begin{array}{c}
X_{cw}/ Z_{cc}\\
Y_{cw}/ Z_{cc}\\
1/Z_{cc}
\end{array}\right]&=H^{-1} \cdot\left[\begin{array}{c}
U_{c} \\
V_{c} \\
1
\end{array}\right],
\label{Zcc}
\end{cases}
\end{aligned}
\end{equation}
where $R_{1},\ R_{2}$ are the first two columns in $R_{3\times3}$.

Finally, with $Z_{cc}$ (depth), we can get the center point $C'$:
\begin{equation}
\label{eq:center}
(X_{cc}, Y_{cc}, Z_{cc}) = Z_{cc}\cdot (X_c, Y_c, 1).
\end{equation}

\textbf{3D Localization} A 3D localization method with a closed-form solution is introduced to calculate head point $H'$ and tail point $T'$, which are on the following two lines respectively:
\begin{equation}
\begin{aligned}
Line_1: (X, Y, Z)&=m \cdot (X_{h}, Y_{h},1), m \in R,\\
Line_2: (X, Y, Z)&=n \cdot (X_{t}, Y_{t},1), n \in R.
\end{aligned}
\label{2 lines}
\end{equation}

\begin{figure}[htpb]
\centering
\includegraphics[width=0.5\textwidth]{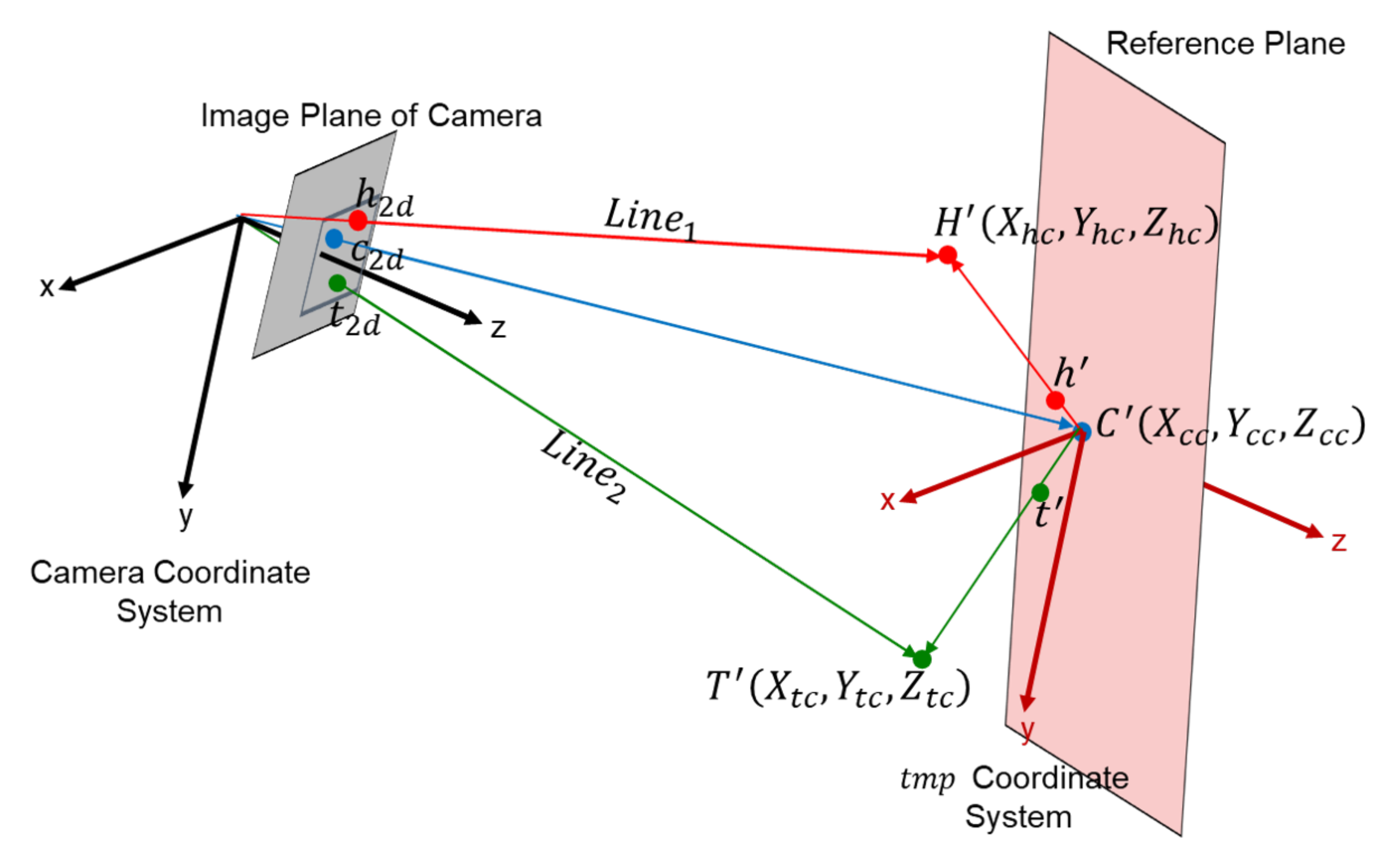}
\caption{3D Localization: $h'$, $C'$, and $t'$ define the relative 3D pose. $H'$, $C'$, and $T'$ defines the absolute 3D pose.}
\label{3D localization}
\end{figure}

As illustrated in Fig.\ref{3D localization}, we set up a new coordinate system with unit being $pixel$, called $tmp$, whose origin point is at $C'$ and has no relative rotation with respect to the camera coordinate system but the translation vector between them is $(X_{cc}, Y_{cc}, Z_{cc})$. In Section 2.2, we got the relative 3D pose, i.e., $h$, $c$, and $t$ in the camera coordinate system with unit being $pixel$. Now we want to estimate the absolute fish pose centered at $C'$ so we translate $h$, $c$ and $t$ to the $tmp$ coordinate system by translating $c$ to $tmp$'s origin point:
\begin{equation}
\begin{aligned}
c'&=(X_{cc}, Y_{cc}, Z_{cc}),\\
h'&=(x_h-x_c, y_h-y_c, z_h-z_c) +  (X_{cc}, Y_{cc}, Z_{cc}),\\
t'&=(x_t-x_c, y_t-y_c, z_t-z_c) +  (X_{cc}, Y_{cc}, Z_{cc}),
\end{aligned}
\end{equation}
where $t'$ $(x_{tc}, y_{tc}, z_{tc})$ and $h'$ $(x_{hc}, y_{hc}, z_{hc})$ are relative head point and tail point in the camera coordinate system. Note that the absolute head point, $H'$, and tail point, $T'$, must also lie on the following lines $C'h'$ and $C't'$ respectively:
\begin{small}
\begin{equation}
\begin{aligned}
(X, Y, Z)&=a \cdot (x_{hc}, y_{hc}, z_{hc}) + (1-a) \cdot (X_{cc}, Y_{cc}, Z_{cc}), a \in R,\\
(X, Y, Z)&=b \cdot (x_{tc}, y_{tc}, z_{tc}) + (1-b) \cdot (X_{cc}, Y_{cc}, Z_{cc}), b \in R.
\end{aligned}
\end{equation}
\end{small}
So, $H'$ must be the intersection point of line $C'h'$ and $Line_1$, and $T'$ must be the intersection point of line $C't'$ and $Line_2$. To find $H'$ and $T'$, we minimize the distance between two lines, which results in a closed-form solution with the simple least-squares method~\cite{charnes1976equivalence}, omitted due to the pages limit.
Finally, and the absolute fish length can thus be calculated:
\begin{equation}
\begin{aligned}
{Length} &= {\mid H'T'\mid}\cdot{\frac{\overset{\frown} {h t}}{\mid ht\mid}},
\end{aligned}
\end{equation}
where $\frac{\overset{\frown} {h t}}{\mid ht\mid}$ is the bending ratio. The left bottom corner of Fig.\ref{pipeline} is a 3D localization example.


\section{Experiments}
\label{sec:pagestyle}
Due to the complexity of collecting data during longline fishing, we have 738 ground-truth labeled fish lengths measurement histogram and the corresponding several hours' stereo videos under different weather conditions and view angles. We can only use the difference between predicted length histogram and ground truth length histogram to evaluate all competing methods. Specifically, we use bias, root mean square deviation (RMSD), Kullback-Leibler (KL) divergence~\cite{kullback1951information}, and earth mover's distance (EMD)~\cite{inproceedings} to access the performance.
For all frames of the same fish in one short video clip, we estimate its length in each frame and use a Gaussian distribution to remove outliers beyond $2\cdot\sigma$ and take average of the rest predicted lengths.

\begin{figure}[htpb]
\centering
\includegraphics[width=0.4\textwidth]{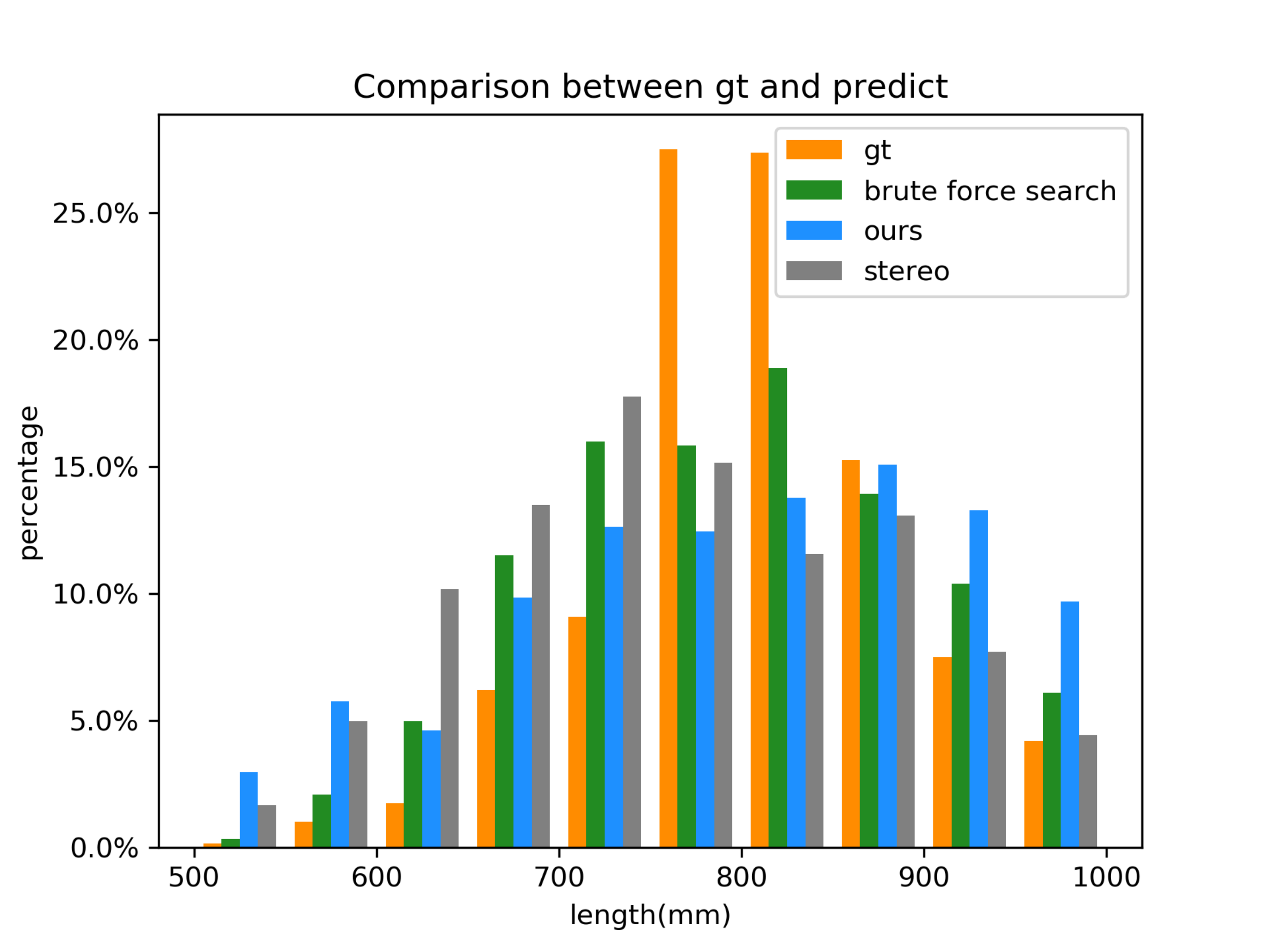}
\caption{Results: Only check lengths in $[500, 1000]mm$ because most lengths are within this range. $y$ axis is percentage of fish number in bins over total number. $x$ axis is length bins.}
\label{comparsion}
\end{figure}

\textbf{Comparison with the state-of-the-art}
Compared with~\cite{huang2018fish}, which requires stereo image pairs as input and is more difficult to deploy in the challenging at-sea environment while ours only needs a single-view image. Fig.\ref{comparsion} and Table.\ref{evaluation} show superior performance of our method in all four metrics.

\begin{table}[htbp]
    \caption{Comparison Evaluation and Ablation Study}\label{evaluation} \centering
    \setlength{\tabcolsep}{1.5mm}{\begin{tabular}{cccccc}
        \toprule[1.5pt]
        Method & Bias(mm) & EMD(mm) & RMSD & KL\\
        \midrule[1pt]
        Stereo~\cite{huang2018fish} & -40.5 & 46.0 & 7.9\% & 0.26\\
        BFS & -10.2 & 24.2 & 5.6\% & 0.11\\
        BFS w/o Bending & -55.4 & 60.0 & 7.9\% & 0.28\\
        Ours w/o Bending & -95.4 & 99.3 & 10.4\% &0.53\\
        Ours & -9.3 & 43.1 & 7.3\% &0.23\\
        \bottomrule[1.5pt]
    \end{tabular}}
\end{table}

\textbf{Ablation Study} To replace our optimization part, we construct a database consisting of thousands of orthographic projection images of relative 3D fish template under various deformation parameters. Given a 2D mask of a target, we brute-force search (BFS) to find one projection image with maximum intersection of union with this 2D mask and use its deformation parameters to generate the relative 3D pose for Stage-3. Fig.\ref{comparsion} and Table.\ref{evaluation} show our optimization has favorable performance as BFS but BFS is more time consuming and memory demanding. Besides, we remove bending ratio and test BFS method and ours. Table.\ref{evaluation} shows bending modeling is critical in fish length measurement.

\section{Conclusions}
\label{sec:typestyle}
We proposed a relative deformable 3D fish template, a chamfer distance based optimization method to predict relative 3D pose from a 2D instance mask, and a 3D localization method with a closed-form solution to estimate absolute 3D fish pose. Our experiments show that our monocular method outperforms the state-of-the-art stereo method, our optimization method has favorable performance as brute-force search and our bending modeling is critical in fish length measurement. 


\vfill\pagebreak
\bibliographystyle{IEEEbib}
\bibliography{refs}

\end{document}